\documentclass[sn-mathphys-num]{sn-jnl}

\usepackage{graphicx}%
\usepackage{multirow}%
\usepackage{amsmath,amssymb,amsfonts}%
\usepackage{amsthm}%
\usepackage{mathrsfs}%
\usepackage[title]{appendix}%
\usepackage{xcolor}%
\usepackage{textcomp}%
\usepackage{manyfoot}%
\usepackage{booktabs}%
\usepackage{algorithm}%
\usepackage{algorithmicx}%
\usepackage{algpseudocode}%
\usepackage{listings}%
\usepackage[acronym]{glossaries}

\newacronym{ir}{IR}{Information Retrieval}
\newacronym{llm}{LLM}{Large Language Model}
\newacronym{llms}{LLMs}{Large Language Models}
\newacronym{lm}{LM}{Language Model}
\newacronym{mom}{MoM}{Machine of Meaning}
\newacronym{cra}{CRA}{Chinese Room Argument}
\newacronym{ai}{AI}{Artificial Intelligence}
\newacronym{ml}{ML}{Machine Learning}
\newacronym{nlp}{NLP}{Natural Language Processing}
\newacronym{mle}{MLE}{Maximum-Likelihood Estimation}
\newacronym{rbm}{RBM}{Restricted Boltzman Machine}

\newtheorem{defn}{Definition}%
\newtheorem{prop}{}

\begin{document}
\title[MACHINES OF MEANING]{MACHINES OF MEANING}
\author*[1]{\fnm{Davide} \sur{Nunes}}\email{davex@ciencias.ulisboa.pt}
\author[1]{\fnm{Luis} \sur{Antunes}}\email{xarax@ciencias.ulisboa.pt}
\affil[1]{\orgdiv{LASIGE}, \orgname{Faculdade de Ciências, Universidade de Lisboa}, \country{Portugal}}

\abstract{ 
    One goal of Artificial Intelligence is to learn meaningful
    representations for natural language expressions, but what this entails is
    not always clear. A variety of new linguistic behaviours present themselves
    embodied as computers, enhanced humans, and collectives with various kinds
    of integration and communication. But to measure and understand the
    behaviours generated by such systems, we must clarify the language we use to
    talk about them. Computational models are often confused with the phenomena
    they try to model and shallow metaphors are used as justifications for (or
    to hype) the success of computational techniques on many tasks related to
    natural language; thus implying their progress toward human-level machine
    intelligence without ever clarifying what that means.  
    
    This paper discusses the challenges in the specification of \textit{machines
    of meaning}, machines capable of acquiring meaningful semantics from natural
    language in order to achieve their goals. We characterize ``meaning'' in a
    computational setting, while highlighting the need for detachment from
    anthropocentrism in the study of the behaviour of \textit{machines of
    meaning}. The pressing need to analyse AI risks and ethics requires a proper
    measurement of its capabilities which cannot be productively
    studied and explained while using ambiguous language. We propose a view of
    \textit{meaning} to facilitate the discourse around approaches such as
    neural language models and help broaden the research perspectives for
    technology that facilitates dialogues between humans and machines.}


\maketitle
\section{Introduction}
Language is used to express our thoughts and communicate about the world. Ever
since the inception of \Gls{ai} as a field, language has been a key
target-subject of research, with the goal of learning meaningful representations
of natural language expressions. We find that recent developments of machine
learning models for natural language processing tasks, while promising, also
lead to a fundamental misunderstanding of the capabilities of these models. For
example, language-model-based approaches are often presented as being capable of
``understanding'' language or capturing the notion of meaning
\cite{benderClimbingNLUMeaning2020, benderDangersStochasticParrots2021}
without ever analysing how model capabilities and behaviour differ from the
human experience. Despite the study of language greatly benefiting from
quantitative methods over the last decades \cite{osgood1952nature}, the field of
\gls{ai} seems to lean towards using the formal and quantitative nature of its
approaches to bypass the problem of what \textit{meaning} is, whilst making use
of shallow metaphors to justify and hype research directions.

To create meaningful representations for natural expressions we need to make
clear how they are meaningful and to whom. We tackle the question of how to
build \textit{machines of meaning}, their limitations, how these can be said to
be capable of understanding, how they deal with \textit{meaning},
\textit{semantics}, and \textit{grounding}. In summary, this paper discusses how
representations of expressions are connected to the world from which they emerge
and what that means for computational approaches to language.

The paper is structured as follows. We start with the connection between
language and behaviour, more precisely, what are meaningful \textit{symbols} and
how these are different from other \textit{signs}. We then discuss how these
symbols are \textit{grounded} and what makes them referential to
\textit{objects} (concrete, metaphorical, or behavioural). The following section
characterizes meaning in the computational setting. We show how concepts
originated in philosophy of language, and quantitative methods in linguistics
found their way to computational semantics, and more recently to \Gls{llms}.
Finally, we highlight two important problems in contemporary approaches to
language modelling that we see as major obstacles to the creation of
\textit{machines of meaning}.

\section{On Symbols} 

\begin{figure}[htbp]
    \centering
    \includegraphics[scale=0.4]{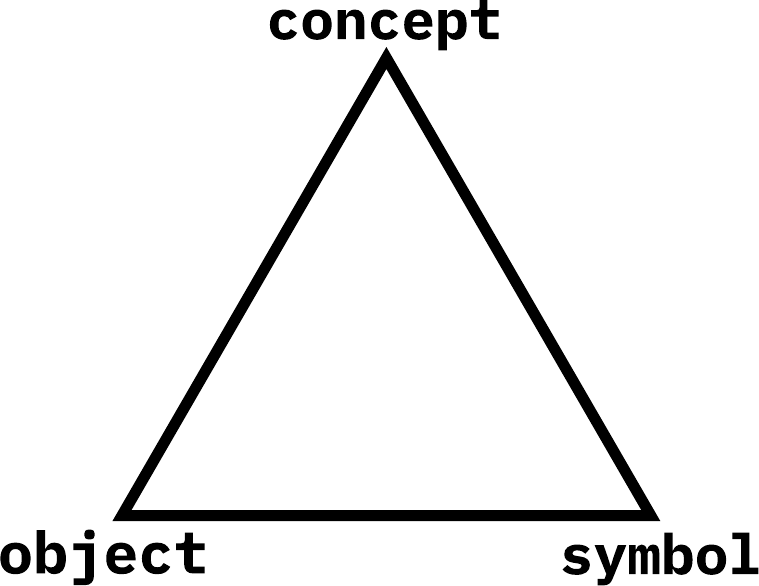}
    
    \caption{Semiotic triad: relates a symbol, a context (object), and a concept
    applicable to the object. Concepts are higher-level categories that might or
    might not apply to a given context or object (e.g. a particular chair versus
    the concept of a chair). Grounding is the process by which symbols are
    related to concepts.}
    \label{fig:semiotics}
\end{figure}

In order to discuss when and how symbols can be about things in the world, and
how we can create machines that establish that connection, we need to clarify
the notion of \textit{symbol}. A long tradition in Semiotics makes the
distinction between a symbol, the object the symbol references, and the concept
associated with the symbol (see figure \ref{fig:semiotics}). For example, we
might talk about the object ``heart'', the concrete human organ that pumps blood
throughout your body, and then the concept of ``heart'' which might be related
to all heart-shaped objects. An object doesn't necessarily need to be a physical
object in the real world, it may also be an abstraction or another symbol.

Each individual connects different objects, contexts, symbols, and concepts for
purposes of communication. This ``semiotic network'' gets dynamically expanded
and reshuffled every time you experience, think, and interact with the world.
For example, when a speaker wants to draw attention to an object or situation,
she can use the concept that applies to the context, then choose the symbol
associated with this concept and render it in speech or some other medium.

The notion of \textit{symbol} itself is involved in a great deal of
misunderstandings, especially when philosophers, cognitive scientists, and AI
scientists discuss the role of symbols in cognition and intelligence
\cite{steels2008symbol}. Cognitive science poses that symbols should be defined
as couplings connecting objects to their categories \cite{clancey1997situated}.
In \gls{ai}, symbols are often defined as \textit{internal representations} with
which computations can be carried out. An extension on De Saussure's work by
Peirce proposes that a \textit{sign} is a combination of
\textit{representament}, \textit{interpretant}, and \textit{object} (the aspects
we used to define semiotic relationships in Figure \ref{fig:semiotics}).
According to Peirce, the \textit{sign} is called a \textit{symbol} if the
connection between these components is arbitrary,  and as a consequence this
relationship must be learned \cite{peirce1974collected}. For example, smoke is
often a \textit{sign} of fire, but certain visual patterns of smoke can turn
into a \textit{symbol} for someone communicating through smoke signals. To the
connection between the symbols and the elicited behaviour, we call
\textit{meaning}.

In what we now call classical \gls{ai}, there was an emphasis on the use of
formal ``symbolic'' representations in rule-based systems in what we called
\textit{cognitive architectures}. Evidence of this conflation of concepts can be
seen in \cite[p. 138]{newell1980}:

\vspace{1em}
\begin{quotation}
\noindent ``Historically, list processing languages have been critically
important in abstracting the concept of symbol processing, and we have
recognized them as carriers of theoretical notions''. %pp(138) 
\end{quotation}
\vspace{1em}

\noindent The additional metaphysical imports assumed by this use of
\textit{symbol} are a source of misunderstandings about the difference between
symbols in computer science (when we talk about symbolic languages) and symbols
as used by cognitive scientists. This view assumes that minds work by
manipulating symbols in the computational sense -- when in fact this is but an
implementation detail. \citeauthor{steels2008symbol} also makes a distinction
between symbols in the linguistic sense and computer science sense and further
recognizes that is probably one of the biggest terminological confusions in the
history of science and ``the debate about the role of symbols in cognition or
intelligence must be decoupled to whether one uses a symbolic programming
language'' \cite{steels2008symbol}.

The misunderstanding about what constitutes a symbol relates to two major
philosophical issues: the \textit{ grounding problem} \cite{harnad1990symbol}
and the \textit{framing problem} \cite{mccarthy1981}. In the \textit{grounding
problem}, especially as it manifests itself in classical \gls{ai}, by
abstracting away certain aspects of the world meant to be perceived by
artificial agents, it is not apparent how symbols can be intrinsic and
meaningful to the agents, since all the grounding is made by a system designer.

In the \textit{framing problem} \cite*{mccarthy1981}, by using a symbol (in the
sense of a symbolic programming language) as an abstraction for a symbol in the
cognitive science sense, we run into the following issue: when using
mathematical logic, how is it possible to write formulae that describe the
effects of actions without having to write numerous accompanying formulae that
describe the mundane, obvious non-effects of those actions? 

The framing problem does have another broader re-interpretation by various
philosophers. For example, in \cite{Dennett1980}, \citeauthor{Dennett1980} poses
the question of how ``a cognitive creature \ldots with many beliefs about the
world'' can update those beliefs when it performs an act so that they remain
``roughly faithful to the world''? This question clashes with the possibility of
belief being itself a linguistic tool that, while it might influence how we
model the world in linguistic terms, has no consequence on how we update our
world models (beliefs included). In fact, in a later essay,
\citeauthor{Dennett1980} acknowledges the appropriation of the \gls{ai}
researcher's view of minds as symbol-manipulating agents with symbolic logic
implementations as abstractions for such symbols \cite{Dennett1986}. The
outstanding philosophical question is how an artificial agent could ever
determine that it had successfully revised all its beliefs to match the
consequences of its actions. \citeauthor{fodor1987frames} suggestively likens
this to “Hamlet's problem: when to stop thinking” \cite[p.140]{fodor1987frames}.
The frame problem, he claims, is ``Hamlet's problem viewed from an engineer's
perspective''. So construed, the obvious way to try to avoid the frame problem
is by appealing to the notion of relevance: only certain properties of a
situation are relevant in the context of any given action, and consideration of
the action's consequences can be conveniently confined to those. The challenge
in building a \Gls{mom} implies one has to deal with a contemporary version of
the framing problem which we will be discussing in our proposed computational
candidate solutions.

\vspace{1em}
\begin{defn}[Symbol]
    \label{def:symbol}
    A symbol is a pattern of behaviour that expresses meaning, but it is not
    meaningful in itself. More precisely, it is the rendering of a social norm,
    being it written (for example a series of characters forming a word),
    spoken, gestured, or another pattern with a semantic role (e.g. smoke
    signals, Morse code, etc).
\end{defn}
\vspace{1em}

Our view is that for a \gls{mom}, symbols themselves do not carry any intrinsic
meaning, the meaning of symbols is always learned and thus part of each symbol
user experience. Taking the definition of symbols given in Definition
\ref{def:symbol}, an artificial \gls{mom} should, in principle, not be
compromised by implementation details. \textit{Meaning} must not be treated as
propositional content of natural language expressions, and should not be
grounded by the system designers. Rather, in a \gls{mom}, a \textit{symbol} is
merely an input or possibly an action in a world of symbol-using agents (human
or computational).

% ***************************************************
% Grounding
% ***************************************************
\section{Grounding}
In 1980, \citeauthor{searle1980} put forward his \Gls{cra} \cite{searle1980} in
order to contradict the notion that intelligent behaviour is the outcome of
purely computational processes in physical symbol systems
\cite{newell1976,newell1980}. Searle's argument touches on two key topics: the
inability of \gls{ai} to deal with symbols that are meaningful to the machine
itself (grounded in its experience); and the difference between programs and
machines (hardware). This argument is therefore at the core of what we call in
\gls{ai}, the \textit{grounding problem}. 

The argument states that only very special kinds of machines can think, with
internal causal powers equivalent to those of the brain, and that \gls{ai} has
little to tell us about thinking since \gls{ai} is ``about programs and no
program by itself is sufficient for thinking.'' therefore, machines cannot know
the meaning of natural language expressions. The argument can be summarized as
follows. 
\vspace{1em}
\begin{quotation}
    \noindent Searle is inside a room and possesses a set of rules (in English)
    that instruct how to produce sequences of Chinese symbols as a response to
    input Chinese symbols. \\\\
    Searle doesn't know Chinese and to him, Chinese
    writing is just a series of meaningless squiggles. \\\\
    Unknown to Searle, people
    outside the room feed him a script, a story, and a set of questions (all in
    Chinese) for which Searle supplies the answers by using the set of rules
    (the program).
\end{quotation}
\vspace{1em}

\noindent Searle assumes that to produce Chinese, he is simply instantiating a
computer program. But from the point of view of someone reading the answers to
the Chinese questions outside the room, the answers seem to be given by someone
fluid in Chinese. This is the scenario under which Searle examines the claims:
\vspace{1em}
\begin{prop}{}\label{prop:understand} The programmed computer understands the
	stories;
\end{prop}

\begin{prop}{}\label{prop:explain} The program in some sense explains human
	understanding. 
\end{prop}
\vspace{1em}
\noindent The first conclusion of the argument is that a computer may make it
appear to understand human language but could not produce real understanding.
Hence, the test is inadequate. Searle argues that the thought experiment
underscores the fact that computers merely use syntactic rules to manipulate
symbol strings, but have no understanding of semantics and suggests that minds
must result from biological processes and computers can at best simulate these
biological processes. While we agree with the part of the criticism, the entire
argument is fundamentally flawed. 

The \gls{cra} raises the question that the understanding of the room is not
grounded in the experience of using Chinese. Perhaps justifiably so (in light of
the technology available at the time), Searle assumes that the rules to
manipulate language are not sufficient for understanding. We object that you can
make that assumption about the system as a whole without specifying where the
rules come from. Let us conceive an alternative scenario for the \gls{cra}.

Suppose we can create a simulated digital replica of a person who knows both
English and Chinese down in such a way that this replica is causally equivalent
to such an individual. Assume this simulation replicates a room and the means to
input and output information to the replica just like in the Chinese room
scenario. Now suppose, that that simulation is written down as a program along
with its state. The person inside the room has access to the state and follows
the rules to input the Chinese symbols into the simulated room. The simulated
room then returns the symbols in English or Chinese, depending on the inputs
(regardless of how laborious this implementation might be for the person doing
the translation between rules and action).

If we accept that causally, a \textit{real} person is the same as a simulated
one, then the man inside the room in the original argument is but a cog in a
larger system, an implementation detail. This has been pointed out in many
replies to the original argument that emphasize a \textit{systems view} of the
problem. If the simulation of a person embodies its exact state, one could
assume that it has an understanding of the symbols it manipulates, and the
meaning of such symbols could be grounded in its experience. For example, Boden
argues that ``Computational psychology does not credit the brain with seeing
bean sprouts or understanding English: intentional states such as these are
properties of people, not of brains'' \cite[p. 244]{boden1988}. In other words,
understanding depends on how the symbol-using \gls{mom} is embedded into the
world, not necessarily related to its implementation details.

% addressing the appeal to biology
According to Searle, an adequate explanation for intentionality can only be a
material one and ``Whatever else intentionality is, it is a biological
phenomenon, and it is as likely to be causally dependent on the specific
biochemistry of its origins''. But Searle's appeal to biology does not imply we
can't study grounding, meaning, and intentionality in artificial settings from a
\textit{systems} perspective, especially because to understand the critical
features of living systems, most biologists accept that a system perspective is
perhaps as fundamental if not more than a material one. For example, a genetic
system can be best understood in terms of information processing and not solely
in terms of its molecular building blocks \cite{smith2000}. What it means is
that in order for a \gls{mom} to be equivalent to a human, we would need to
build an analogue to a human existence down to all details relevant to
characterize human experience. 

%KEY POINT
A similar resistance has been expressed by both biologists and philosophers in
accepting the \textit{systems} perspective: the idea that the same process can
be instantiated (embodied) in different media. For example, in
\cite{penrose1989}, \citeauthor{penrose1989} has argued that intelligence and
consciousness are due to certain (unknown) quantum processes unique to the brain
and therefore any other type of implementation can never obtain the same
functionality. \citeauthor{boden1988} also argues that Searle mistakenly
supposes programs are pure syntax. But the inherent procedural consequences of
programs that bring about the activity of certain machines give them a toehold
in semantics, where the ``semantics in question is not denotational, but
causal'' \cite[p. 250]{boden1988}. Thus, a robot might have causal powers that
enable it to refer to a hamburger. This is similar to the view of
\citeauthor{steels2008symbol} when he proposes that the symbol grounding problem
has been solved.

Chalmers offers a parody in which it is reasoned that recipes are syntactic,
syntax is not sufficient for crumbliness, cakes are crumbly, so implementation
of a recipe is not sufficient for making a cake. The implementation makes all
the difference; an abstract entity (recipe, program) determines the causal
powers of a physical system embedded in the larger causal nexus of the world
\cite{chalmers1996conscious}. 

%OUR VIEW
Our position is that the issue is not whether programs can understand, but which
kinds of programs can understand and in what ways. If we think (like Searl did)
that it is not reasonable to attribute understanding based on the behaviour
exhibited by the Chinese Room, then it would not be reasonable to attribute
understanding to humans based on similar behavioural evidence.
\citeauthor{searle1980} argues that there is an important distinction between
\textit{simulation} and \textit{duplication}. No one would mistake a computer
simulation of the weather for weather, or a computer simulation of digestion for
real digestion. Searle concludes that it is just as serious a mistake to confuse
a computer simulation of understanding with understanding. The problem with this
conclusion is that it is not clear that the distinction can always be made. For
example, are artificial hearts simulations of hearts? Or arse they functional
duplicates of hearts, hearts made from different materials? Do people with
artificial limbs walk? Or do they simulate walking? If the properties required
for something to be a certain kind of thing are high-level properties, the
lower-level properties should be irrelevant for something to be that kind of
thing. Furthermore, the entire field of \gls{ai} has been mostly dedicated to
the understanding and replication of high-level properties of intelligent agents
(human or otherwise) through computational implementations. Chalmers offers a
principle governing when a simulation is replication contra Searle and
\citeauthor{harnad1989minds} \cite{harnad1989minds}: 
\vspace{1em}
\begin{quotation}
    \noindent A simulation of X can be an X when the property of being an X depends only
    on the functional organization of the underlying system, and not on any
    other details \cite[p. 328]{chalmers1996conscious}. 
\end{quotation} 
\vspace{1em}
% in summary (cra)
In summary, and contrary to the position in the \gls{cra}, the capacity of a
\Gls{mom} to understand the meaning of the symbols it processes should not be
dependent on the intentionality of its components. Moreover, displaying an
apparent capacity to understand a language is not sufficient to understand that
language. It is impossible to settle these questions without employing a
definition of the term \textit{understand} that can provide a test for judging
whether the hypothesis of whether a program can understand is true or false
\cite{simonEisenstadt2002}. Such a judgment hinges on a process and core
challenge for \gls{ai} dealing with natural language: the problem of
\textit{grounding}.

% The Grounding Problem: Definition

The \textit{grounding problem} generally refers to how artificial agents can be
embedded in their environment such that their behaviour can be intrinsic and
meaningful to the agent itself, rather than dependent on an external designer
\cite{harnad1990symbol,sharkey1994,riegler1999rethinking,vogt2002physical}.
In the context of Artificial Intelligence, the problem can be stated as: 
\vspace{1em}
\begin{quotation}
    \noindent How does one connect computational representations with what they
represent (e.g. symbols or abstract concepts)?
\end{quotation}
\vspace{1em}
In a computational setting, the \textit{symbol grounding problem} is articulated
in \cite{harnad1990symbol} as analogous to trying to learn Chinese, from a
Chinese/Chinese dictionary alone. Harnad extended and refined Searle's analysis
of the problem in the \gls{cra} and proposes a possible solution. The important
conclusion is that ``cognition cannot be just symbol manipulation''. The symbols
in the Chinese room might very well be interpretable, and have meaning ``but the
interpretation will not be intrinsic to the symbol system itself: It will be
parasitic on the fact that the symbols have meaning for us (the observers), in
the same way that the meaning of the symbols in a book is not intrinsic, but
derives from the meaning in our heads'' \cite{harnad1990symbol}.

In the early seventies, AI experiments like SRI's Shakey robot
\cite{nilsson1984Shakey} could be said to be embodied and embedded in the world
using sensors, actuators, etc, and capable of storing and updating semantic
networks of symbols that could be used to communicate about its state, make
plans, and execute them. Shakey had to perceive the world, construct a world
model, parse sentences and interpret them in terms of this world model, and then
make a plan and execute it. But the \textit{grounding} and the semantic
relationships between symbols were carefully mapped out by human programmers,
the semantics came from us, humans.

\begin{figure}[htbp!]
    \centering
    \includegraphics[scale=0.8]{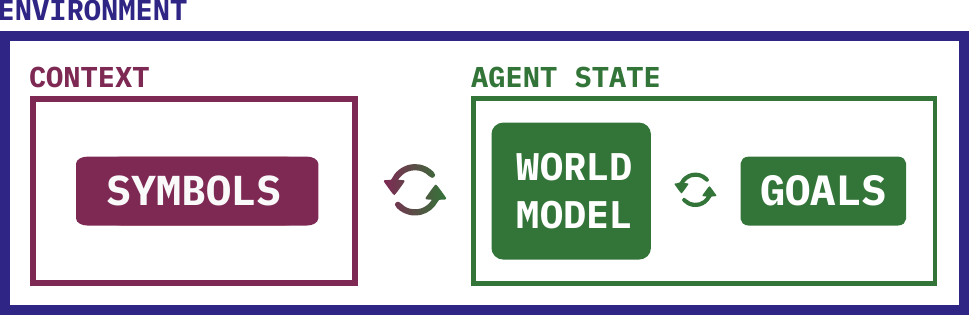}
    \caption{Symbols shape our linguistic models of the world. The process of
    grounding of symbols is done by interacting with other members of a
    symbol-using community (e.g. human speakers of a particular language, or
    other \textit{Machines of Meaning}), and by experiencing the contexts in the
    world, where symbol use is relevant.}
    \label{fig:grounding}
\end{figure}

\noindent A \gls{mom} must be able to learn and use the relationships between symbols and
what they reference autonomously, grounding their semantics in experience guided
by their motivations (whether these motivations are acquired or designed)
(see figure \ref{fig:grounding}) \cite{antunesnunes2014}. Grounding can be
defined as follows.

\vspace{1em}
\begin{defn}[Grounding]
    \label{def:grounding}
    The grounding of a \textit{symbol} refers to the process whereby an agent
    updates its \textit{world model} by experiencing the contexts where the use
    of that \textit{symbol} is relevant and relevance is guided by the agent's
    \textit{goals}.
\end{defn}
\vspace{1em}
\noindent There are two main points on which most approaches to grounding seem
to agree: 

\begin{enumerate}
    \item Solving \textit{grounding problem} is fundamental to the development
    of intelligent behaviour
    \cite{brooks1991a,brooks1991b,sharkey1994,jackson1996}
  
    \item To achieve grounding, the agents need to be ``hooked''
    \cite{jackson1996} to the external world: That means there has to be
    \textit{causal} connections, which allow the artificial agent's internal
    mechanisms to interact with their environment directly and without being
    mediated by an external observer. 
\end{enumerate}

\noindent From this understanding of grounding, it follows that
\textit{grounding} and \textit{meaning} are dependent on how symbol users are
embedded in the world. In the case of human beings, the way our cognition is
embodied determines not only how language emerges but also its semantics; not
only because there are no causal distinctions between mind and body, but because
our self-awareness makes the body part of the environment. For example, having a
digestive system allows us to ground the word ``hunger''. Hunger is a source of
beliefs, a source of drives/motivations, it provides an input for the reasoning
systems, it forms an output, in fact, a cognitive output of a very physical
experience of digesting food. Consider also how a conversation between a human
and an artificial system could resolve the gaps provided by each other's
representation of the originating signals (some of which can be cognitively
acknowledged as symbols) for the agent's consciousness/awareness system.

An artificial \gls{mom} can, in principle, even as a \textit{disembodied}
entity, acquire analogous concepts but only from a behaviourist perspective. The
grounding process is limited by how the words it uses are relevant to the
machine. A robot will never be ``hungry'' in the way a human is, but it can
ground the same word if it's relevant to its goals --for example to feed
``hungry'' humans. This also means that first-hand experience is not relevant
for the grounding of linguistic expressions, but a common framework of
embodiment is still crucial for communication since our experiences (and thus
our grounding processes) are more or less uniform. 

% **************************************
% ON MEANING
% **************************************
\section{The Meaning of Meaning}
In recent advances in \gls{ai} in the domain of natural language, the word
\textit{meaning} is treated superficially as if no further clarification is
necessary. Often, such advances are contrasted with human performance in various
benchmarks and make vague claims about language \textit{understanding} or
\textit{representation} without elaborating on what this entails. This
anthropocentric but superficial discourse surrounding \gls{ai} for natural
language creates misconceptions, unwarranted hype, and further contributes to
stifling possible research directions. We shall try to clarify the notion of
\textit{meaning} in a way that helps with creating a definition for machines of
meaning. We will discuss two perspectives on the notion of \textit{meaning} that
are frequently conflated. Furthermore, we can look at meaning in terms of its
\textit{origin}, how does meaning come about; and its \textit{ontological
nature}, or what exactly is meaning. 

Wittgenstein's work \cite{wittgensteinTractatus} and the one of Saussure
\cite{saussure1916course} are seminal in tackling many problems surrounding
meaning, problems which were inherited from $19^{th}$ century Philosophy and
Linguistics. At the turn of the $20^{th}$ century, both Saussure and
Wittgenstein departed radically from the practices of their disciplines by
doubting whether the questions posed by their predecessors and contemporaries
made any sense at all. One such misconception was that words are names of
objects or properties existing prior to language. This kind of surrogate view of
language can be traced back to Plato's \textit{Cratylus}, as well as Leibniz and
Locke. Simply put, this was the view that words have meanings because they stand
for something extra-linguistic \cite{harris1990language}.

The answers found by both Wittgenstein and Saussure were similar in that both
rejected that \textit{meaning} could be isolated from the rest of the language
system
(or from the language speakers). To understand meaning was to understand its
link to language use. Both authors saw words as arbitrary symbols, and in order
for some communication to occur, some restrictions had to guide this
arbitrariness. Vocabulary, semantic relationships, and grammar were seen as
restrictions that would help settle this arbitrary nature of words into norms.
With this came one of the most cited views over semantics, widely adopted by
computational semantics, and more recently by machine learning approaches for
language modelling or natural language processing: 
\vspace{1em}
\begin{quotation}
    \noindent For a large class of cases though not for all in which we employ
    the word ``meaning'' it can be defined thus: the meaning of a word is its use
    in the language. \cite[Sect. 43]{wittgenstein2010philosophical}.
\end{quotation}
\vspace{1em}
The \textit{meaning as use} perspective, does not paint the entire picture, but
it does make some aspects clear: how symbols acquire their meanings is dependent
on multiple social, cultural, and individual factors. Meaning has a purpose: to
communicate about the world in such a way that is relevant to the community
where it emerges. Creating and maintaining coherent semantic networks can be
seen as a cooperation problem \cite{lewis2008conventions}. The meaning
assignments are arbitrary, but the substrate in which they emerge is not, and
humans are known to be very good at exploiting useful patterns to understand and
communicate about complex contexts --as it's evident from our use of
\textit{folk psychology} to understand and reason about the minds of others
\cite{dennett1991real}.

Wittgenstein does not tell us exactly what meaning \textit{is} (the ontological
side of the problem), but his functional and structural perspective over
language, semantics, and \textit{meaning}, paved the way for two major
perspectives on how to reason about \textit{meaning}: \textit{logic positivism}
and \textit{scepticism}. The former asserted that meaningfulness arises from the
empirical observation of logical proof. \textit{Scepticism} challenges the idea
that expressions of our language possess such a thing as determinate meaning:
ascriptions of meaning are themselves meaningless and thus unjustifiable from a
purely formal point of view: understanding and knowledge are private, the fruit
of an individual's experiences. 

The opposing views shared important points, most notably in methodology. The
``linguistic turn'' of which Wittgenstein was part has led to the conclusion
that many traditional philosophical problems arise when we become confused by
words used unusually so that we do not understand what they mean in a certain
context. In Peter Hacker's criticism of the neuroscientific terminology
\cite{bennett2003philosophical}, there are several examples of this
problem. For instance, if we become puzzled by questions like ``Can
consciousness be located in the brain?'', ``Is consciousness essentially
private?'', first we have to work out what those questions mean, because
consciousness is not just a technical or well-defined term, nor is it only used
in uncontentious situations without any philosophical relevance.

To characterize \textit{machines of meaning}, and to build them as artificial
agents embedded in our linguistic worlds we take the position of
\textit{sceptics} in that any ascriptions of meaning to symbols are arbitrary,
socially constructed, and thus, themselves normative which implies that
\textit{meaning is not content}. It is private in the same sense that norms are
private, but socially constructed and reality-involving. This contrasts with
earlier views of Russell and Frege on the possibility of the existence of a kind
of perfect logic language that would be more suitable to explore the
philosophical problems of language because the imperfections in natural language
made them inadequate for their philosophical role
\cite{russell2009philosophy,wittgensteinTractatus,frege1948sense}. Similarly,
Jerry Fodor independently proposes a well-known empirical hypothesis, of a
\textit{language of thought} as an integral part in its representational theory
of mind \cite{fodor1975language}. Roughly, the hypothesis says that thinking is
expressed in a symbolic system realized in the brain. This symbolic system has a
syntax and truth-conditional semantics. The hypothesis has informed numerous
studies in the cognitive sciences \cite{Bianchi2020}. This more ``mentalistic''
approach to language can also be found in the work of Chomsky on language
acquisition \cite{waller1977chomsky}. 

We contend that an excessively formal account of language can lead us to
confound a model of a phenomenon with the actual phenomenon and a clear example
of this confusion can be seen as the consequence of what Wittgenstein describes
as the rule-following paradox
\cite{feyerabend1955wittgenstein,kripke1982wittgenstein}:
\vspace{1em}
\begin{quotation}
    \noindent this was our paradox: no course of action could be determined by a
    rule, because every course of action can be made to accord with the rule.
\end{quotation}
\vspace{1em}
This relates to meaning and mental content in the sense that meaning something
by a linguistic expression can be seen as analogous to following a rule. If the
meaning we assign to a word is expressed in terms of a certain behaviour, then
there must be a fact in virtue of which we mean to use a certain word following
a certain expectation or rule, and if there's such a fact, then we can model our
meaning in terms of that fact, and use the rule to express that meaning. The
paradox implies that there is no such fact.  

\citeauthor{kripke1982wittgenstein} develops this into a more explicit
formulation \cite{kripke1982wittgenstein}. Suppose that \textit{plus} and
\textit{quus} are mathematical functions, which we can symbolize as $+$ and
$\oplus$ respectively. Plus, or the addition function, is defined in the usual
way. The \textit{quus} function is the same as the addition or plus function for
certain inputs but diverges past a certain point. More specifically, it is
defined as follows:

\begin{equation*}
    x \oplus y = x+y if x, y < 57, 
    x \oplus y = 5 \text{ otherwise}
\end{equation*}
\\
\noindent The \textit{sceptical} question is this: in the past, by the symbol
$+$, did you mean \textit{plus} or \textit{quus}? The \textit{sceptic} argues
that it seems that there is no fact of the matter about whether one means
\textit{plus} or \textit{quus}. But if this is the case, then presumably the
problem is general, and there are no facts about meaning at all. The underlying
hope in using rules to subsume meaning is that one could reason about semantics
using the precise mathematical framework of logic, but the paradox shows us that
rules are not self-contained or independent of external factors, which makes
this an inadequate model --\textit{meaning} cannot, therefore, be modelled as
propositional content or a set of truth values because this leads to the
\textit{grounding problem}.

A solution to the rule-following paradox comes from the normative nature of
language, in that the kind of norms at stake are grounded in meaning, rather
than grounding meaning. Linguistic norms arise from the inherent social function
of language. We desire to communicate information. Such communication is in
practice only feasible if the participants use words consistently. This leads to
the conclusion that interactions between individuals contribute to the
convergence of individual linguistic world models. This idea is present in later
Wittgenstein investigations expressed in the form of cooperation through
``language games'' \cite{feyerabend1955wittgenstein} and in
\cite{lewis2008conventions} where we can look at the construction of meaning and
norms through the lenses of \textit{coordination problems} in general, and game
theory in particular. The language one uses depends on the conventions we are
party to. These conventions are regularities in behaviour patterns, sustained by
an interest in coordination and an expectation that others will do their part
\cite{lewis2008conventions}.

There may be no facts in virtue of which we can justify what is true about the
meanings we use, but we can analyse linguistic behaviour and language use in a
larger normative context. Behaviour is not born out of rules but the opposite,
rules are models for consistencies observed in behaviour. In fact, these
consistencies in observed behaviour serve as the basis for determining whether
an agent is semantically making use of symbols. As an example, it was found that
vervet monkeys are capable of acquiring semantics while using a fixed
vocabulary, in other words, they are capable of attributing and maintaining
normative meaning to the \textit{symbols} they use and experience in their
communities \cite{seyfarth1980vervet}. We know they learn the association
between calling signs and their meanings because their young are born with the
capacity for emitting a fixed set of call signs, but they are not born with the
instinct for when these call signs should be used; instead, they use them
erroneously at first and learn to use them correctly by observing adult
behaviour. 

We must look at \textit{meaning} not as an object or a state, but as a highly
contextual process. For a \textit{machine of meaning} (humans included) it
matters not when a sentence is true, it only matters that a sentence is
consistent with the machine's goals and motivations. Wittgenstein expresses this
idea concisely in Philosophical Investigations \cite{feyerabend1955wittgenstein}
as:
\vspace{1em}
\begin{quotation}
    \noindent What one wishes to say is: ``Every sign is capable of
    interpretation, but the meaning mustn't be capable of interpretation. It is
    the last interpretation.''
\end{quotation}
\vspace{1em}
Independence between \textit{symbol} and interpretation also implies that
meaning is a reality involving process that leads to normative behaviour, not a
collection of properties and predicates about said reality, and while we can
create logic-based models that express truths about norms, it doesn't follow
that there are facts and truth-values about the said norms outside that model,
nor does it imply that such models could be universals we could use to talk
about all meaning. If we managed to create such a model, then we could probably
call it a theory of meaning, but the theories applicable to specific instances
of machines of meaning (e.g. computational implementations) should not extend to
other instances as universals (e.g. to human notions of meaning). So, rather
than considering this a chasm between psychological and logical philosophical
approaches, it is more constructive to characterize \textit{meaning} without
resorting to some ideal framework.
% END rule-following paradox

In ``The Meaning of Meaning'', \citeauthor*{putnam1975meaning}
\cite{putnam1975meaning} tackles what has been labelled \textit{semantic
externalism}: the view that what is going on in an individual's mind is not
(entirely) determined by what is going on inside their bodies. In Putnam's famed
\textit{Twin Earth} thought experiment, elsewhere in the universe, there is a
planet exactly like Earth in virtually all aspects, a \textit{Twin Earth}. The
one difference between the two planets is that there is no water on Twin Earth.
In its place, there is a liquid that is superficially identical, but chemically
different. This argument attacks the notion of determination by imagining a
context in which we should find it is natural to say that although speakers on
Earth and \textit{Twin Earth} express the same intention when they use
\textit{water}, the extension of that term --and hence its meaning-- might, by
our current lights, differ according to the community's environment. The issue
with such a scenario is that it's designed as what Daniel Dennet calls an
\textit{intuition pump}, in that it guides the thinker through the problem by
making use of their intuition while turning a blind eye to certain aspects of
the thought experiment: it \textit{pumps} the thinker's intuition in the desired
direction. The \textit{Twin Earth} directs our intuition towards accepting that
references and their extensions are not dictated solely by what is in our heads
as the objects outside result in the same kind of subjective experience while
being objectively different when analysed under a different context (e.g a
chemical analysis of the \textit{water}). This is misleading because if a need
for a distinction between waters did arise (e.g. if someone could travel between
them), then the name \textit{water} would probably diverge between Earths. Yes,
the meaning of \textit{water} is different between \textit{Earth} and
\textit{Twin Earth}, but the real question is to whom.

Semantic externalism does help us highlight some aspects of \textit{meaning}.
Namely, that that individual psychological states do not determine extensions;
that an individual in isolation cannot in principle grasp any concept; that an
individual's grasp of concepts does not determine the extension of all the
individual's terms; and meanings are best not conceived as entities or objects
at all. Semantic externalism also poses that: first, meaning is
reality-involving in the sense that it's significantly determined by reference
rather than vice-versa; secondly, the grasp of meaning is essentially social in
character; finally, individual notions of meaning, concepts, and beliefs, ought
to be settled by a range of cultural, environment-involving factors, including
purposes and context(s) of a speaker's assertion, the use of conventions within
a linguistic community, and ultimately shaped by each individual's
self-interests. This summarizes the two senses in which we can refer to
\textit{meaning}. The first is a process directed by each individual's
motivations and their relationship with their environment. The second is a
designation for normative behaviour observed in a given community.

Computationally, we can see meaning as the multitude of states a
machine can go through when primed by language in a given context. It does not
imply that we should treat such states as meaningful object representations. A
\gls{mom} should be able to adapt in the process of what Wittgenstein calls
\textit{language games}: a series of coordination events where one integrates
feedback about symbol use \cite{wittgensteinTractatus}. There is plenty of
evidence for this continuous adaptation in humans. For example, from Psychology,
studies in natural dialogue \cite{pickering_garrod_2004} show that speakers and
hearers align their communication systems at multiple levels within the course
of a single conversation. From linguistics, we see evidence showing how we
engage in intense problem-solving to solve inconsistencies in our grammar and
align our linguistic systems with each other \cite{francis2003mismatch}.

The connection between individual and collective goals and norms can be seen in
something like colour names. For example, humans with standard vision can see
millions of distinct colours, but we use a small set of words to describe them.
In English, most people get by with 11 colour words: black, white, red, green,
yellow, blue, brown, orange, pink, purple and grey. In a specialized activity,
we might make use of as many as 50 or 100 different words for colours, but this
is still a tiny fraction of the colours that we can distinguish. Interestingly,
the number of symbols we use to describe colours varies greatly and depends on
cultural and environmental aspects, for example, on whether a language is
developed in industrialized societies. Bolivian Amazonian language
\textit{Tsimane} has only three words that everyone knows, corresponding to
black, white, and red. The work in \cite{gibson2017color} found that in all
languages warm colours are easier to communicate and distinguish than cool
colours, this occurs despite some languages having significantly fewer terms
that people use consistently than others. Furthermore, the distinction between
colours and the number of words we use to talk about them is correlated with the
number of things we want to talk about or rather that are useful for a given
community to talk about. This also explains the difference between
industrialized and non-industrialised societies.

In a tentative way to identify the relevant aspects of meaning  that should be
taken into account when building \gls{mom} and summarize them as follows:

\begin{defn}[Meaning]
    \label{def:meaning}
    Meaning is the learned connection between a symbol and its referent in a
    context. A context is a set of features, or consistent behaviour. What makes
    the features of a certain linguistic context meaningful is its importance
    for a certain goal such as survival or coordination. 
\end{defn}

%*************************************
%   MACHINES OF MEANING
%*************************************

\section{Machines of Meaning}
Many claims about computational model capabilities are often made through the
use of words like \textit{understanding} or \textit{meaning} as if the use of
such expressions requires no further clarification. In the previous sections we
discussed and clarified these concepts and highlighted how their use can be
problematic. We now discuss how existing concrete technical implementations of
computational models are related to the philosophical concepts previously
mentioned. Taking modern \Gls{lm} implementations as a starting point, we will
discuss their limitations, and propose solutions to make such models truly
embeddable in our linguistic world. We believe modern implementations of
language models are already \textit{machines of meaning} but misconceptions
about \textit{meaning} lead to a failed alignment of expectations between what
these models can do and what we think they ought to be capable of.  This is made
evident for instance in the work of
\citeauthor{benderDangersStochasticParrots2021}, where language model
capabilities are criticized as a way to make a case against the hype surrounding
such technologies, especially when this unfounded optimism influences not only
the research landscape, but also policymaking regarding the use of language
modelling at scale
\cite{benderClimbingNLUMeaning2020,benderDangersStochasticParrots2021}.

% work related to LM and to ``artificial'' machines of meaning
\subsection{From Structuralism to Multimodal Language Modelling}
Wittgenstein and De Saussure established a new perspective for the study of
meaning and for later developments in structuralism. Notably, Saussure
introduced the basics of several dimensions under which language could be
systematically studied, namely, \textit{syntagmatic} and \textit{paradigmatic}
analysis, under which elements (e.g. words) in language are studied according to
their contrast with other elements either from a syntactic or a lexical
perspective, respectively \cite{saussure1916course}. This analytical framework
is mostly concerned with the abstract structural patterns (phonetic,
morphological, syntactic, or semantic) found in language, and later developments
such as the work of \citeauthor{harris1952discourse} are mostly detached from
the psychology of language, and instead focused on mathematical techniques for
tackling these structural elements \cite{harris1952discourse}.

%In contrast, his student Chomsky reacted strongly against indifference toward
%the mind, and viewed structural elements as individual, non-social, and
%internalized mechanisms.

\noindent This analytical view over language presents a first bridge between
linguistics, philosophy of language, and modern computational linguistics. One
of the defining features of this line of work is the definition of
distributional syntactic categories. For example, what makes \textit{cat} and
\textit{apple} nouns is the fact that they can appear in the same position in
sentences. \citeauthor{harris1952discourse} viewed the relations between
elements in a syntactic structure as probabilistic relations such that the
occurrence of one increases the probability that the other will occur.
Structural linguistics finds its way into computational methods and in
particular into \gls{ir} systems, in the form of the \textit{distributional
hypothesis} \cite{harris1954distributional}:
\vspace{1em}
\begin{defn}[Distributional hypothesis]
    \label{def:distributional}
    Words that occur in the same contexts tend to have similar meanings.
\end{defn}
\vspace{1em}
\noindent The underlying idea that ``a word is characterized by the company it
keeps'' was also popularized by Firth \cite{firth1957} and is present in the
work of Weaver in a memorandum on the possibility of using computers to solve
the world-wide translation problem \cite{weaver1952translation}. This
characterizes semantics as the quantitative analyses of data, where text corpus
(a large set of texts) can be used to build statistical models of word
co-occurrence.  The work in \cite{sahlgren2008} discusses the distributional
hypothesis and illustrates how this simple idea morphed into modern approaches
for statistical semantics, and, more importantly, how a statistical analysis of
structural elements in language turned into a problem of
\textit{representation}. 

Statistical Semantics (\cite{weaver1952translation}) is the study of how the
statistical patterns of human word usage can be used to figure out what people
mean, at least to a level sufficient for information access
\cite{landauer1984statistical}. Our contention is in the fact that this informal
``disclaimer'' of ``at least to a level sufficient for information access'' is
often overlooked, or completely ignored to make unsubstantiated claims about the
capacity of computational models to deal with \textit{meaning} or, more
precisely, to capture the notion of meaning. For example ``a representation that
captures much of how words are used in natural context will capture much of what
we mean by meaning'' can be seen in \cite{landauer1997solution}. This
characterization of meaning is insufficient, for the reasons we previously
discussed, but nevertheless is prevalent in most literature today. 

% core statement 
The implication from Wittgenstein that ``meaning is use'' that we see
transported to statistical semantics as a way to acquire and represent knowledge
in \cite{landauer1997solution} does not imply that the nature of meaning lies
solely in the patterns of language use. What it implies is that whatever meaning
is, it is established by our use of language; and because language use is what
defines our arbitrary associations between symbols and the expectations of
language speakers, naturally, the patterns of language use will reflect aspects
of the world. It does not follow that if we can formally represent language-use
patterns, we can extract meaning from them --in the same way that we cannot get
meaning out of a snapshot of the state of a human body, or from a set of rules
made to model language use.

% What is information retrieval and how is connected to representation
The distributional hypothesis and the algebraic methods developed from it found
their way into \Gls{ir}, to find material (usually documents) of an unstructured
nature (usually text) that satisfies an information need from within large
collections (usually stored in computers). The idea of being able to acquire and
represent meaning is appealing because if we can manipulate formal meaningful
representations for natural language expressions, with the help of computational
methods, we would be able to organize, search for, and retrieve vast amounts of
information in an instant. We would be able to turn a resource like the World
Wide Web into what would be in effect, a semantic database. The original idea
for a \textit{semantic web} as envisioned by Tim Berners-Lee as an extension of
the World Wide Web through standards like formal ontologies
\cite{berners1999weaving} fell short of its goals precisely because of the same
issues we discussed before related to symbols, and grounding. Ontologies are
supposed to be formal representations of concepts of which the corresponding
words in textual documents are instances; this is, much like what happened in
early \gls{ai} approaches, the connection between symbol and concept is seen as
straightforward, but ontologies are not grounded in computational agents, they
are grounded in humans that make use of and agree upon their structure and
lexicon. %According to one celebrated anecdote, semanticists can model the
%meaning of life once and for all just as \textit{life\'}.

\subsection{From Structuralism to Large Language Models}
It is easy to see the appeal of computational methods to investigate the
\textit{distributional hypothesis}, nevertheless, in the literature on language
modelling, there is often a conceptual jump from (1) \textit{modelling the
structural patterns of language} to (2) \textit{we can represent meaning because
meaning is use}. Nevertheless, we can still find in each methodological progress
of computational methods, steps towards a feasible version of what we could
consider an artificial \Gls{mom}. 

The foundations of language modelling are laid out in \cite{brown1992class},
where the production of text in a given language (in this case English) is
modelled by a set of conditional probabilities, what we call \textit{n-gram}
models. The probability of a given word to occur in a language is characterized
as a product of conditional probabilities of the previous words. To the
computational mechanism for obtaining these conditional probabilities, we call a
language model. One of the problems with this kind of models lies in the fact
that vocabularies are generally very large, so even for small sequences of
words, the number of parameters in a model would be huge because we need to keep
track of the probability of all words to co-occur with all contexts. An
\textit{n-gram} model is also incapable of dealing with unknown words, it can
only parse predetermined vocabularies. Moreover, data sparsity means that some
contexts may not occur frequently or at all in the training data, resulting in
low or zero probabilities. Finally, it's difficult to incorporate the notion of
grounding in these models since their objective is solely to maximize how
accurate probabilities of words occurring in training data are. The
representations of words (their occurrence counts) are predefined and cannot
account for other kinds of semantics (how these words are used by other agents
in the world).

The problem of sparsity can also be seen as an induction problem and a tentative
solution is given in the work related to \textit{Latent Semantic
Analysis/Indexing} \cite{deerwester1990indexing,landauer1997solution}. In a
nutshell, the authors propose that dimensional reduction can help as a way to
tackle the previous issue of data sparsity and simultaneously, as a solution to
the induction problem: finding a mechanism by which a set of stimuli, words, or
concepts come to be treated the same for some purposes. Techniques from
numerical linear algebra like Singular Value Decomposition (SVD)
\cite{shlens2014tutorial} allow us to analyse the term-term or term-document
co-occurrence matrices, and to construct low-rank approximations of these
matrices. We can reduce the dimensionality of the features that define each word
(its co-occurrence in numerous contexts) while preserving as much information as
possible. We can construct representations where the similarity between
statistical patterns of usage is not solely dictated by the specific
observations from training data and thus generalize to unseen data. The authors
argue that this approach displays ``human-like generalization'' that is based on
learning and doesn't rely on primitive perceptual or conceptual relations or
representations''. They further argue that perhaps this is a step in solving the
mystery of how humans are capable of generalization from so little experience
like the case of language acquisition \cite{chomsky1991linguistics}.

% 2. Neural language modelling The idea of a distributed representation 
Distributed representations in the form of neural network models solve a great
deal of problems we find with the previous numerical methods of representation
learning. From a neurocognitive science perspective, there have been many
proposals about how concepts might be represented in biological neural networks:
from a single neural unit representation where each concept is represented by a
single neuron \cite{barlow1972single} to distributed representations where each
concept is represented as an activation pattern over the cortex
\cite{lambon2014neurocognitive}. As we discussed previously and has
\citeauthor{hinton1986learning} point out in \cite{hinton1986learning}, it's not
clear how the one-to-one representations can be implemented without some bias
from the system designer. Distributed representations implemented in artificial
neural networks completely bypass this issue and offer us the ability to
generalize for unseen data: when parameters are updated in an artificial neural
network model to incorporate an experience about a particular symbol, the model
automatically updates its state to take into account similar contexts (that
might not even occur in seen data).

In \cite{schmidhuber1992learning}, \citeauthor{schmidhuber1992learning}
highlights one of the principles behind most modern representation learning
approaches including neural-based language models. The work in
\cite{schmidhuber1996sequential} also highlights another important concept of
language modelling using artificial neural networks, and distributed
representation learning in general: the fact that they are essentially methods
of compression using prediction. If a model can learn to predict sequences of
symbols, we can use it as a way to compress those sequences of symbols (like
documents), by leveraging structural properties of language to memorize large
amounts of data. For large enough datasets, we could even think about \gls{llm}
as a lossy version of compressed textual data from all over the Web, since these
are trained to maximize how well they can predict textual sequences but also to
satisfy the human text prompts. This doesn't mean they can or should be seen as
reliable information retrieval systems
\cite{benderDangersStochasticParrots2021}.

Perhaps one of the most seminal examples of the usage of neural networks to
model language, and by extension the structural properties of language, is the
work on neural probabilistic language models \cite{bengio2000neural}. This
approach to the modelling of sequences and co-occurrence patterns overcomes the
previous issues of numerical-based dimensionality reduction methods and learns
the joint probability function of sequences of words in a language. The model
learns simultaneously (1) a distributed representation for each word and (2) the
probability function for word sequences, expressed in terms of these
representations. Generalization is obtained automatically, because a sequence of
words never seen before is assigned a high probability if it is made of words
that are similar (in the sense of having a similar representation) to words
forming an already-seen sentence. Learning a distributed representation for
words allows each training sentence to inform the model about an exponential
number of semantically neighbouring sentences. This approach is a more promising
candidate to be a \gls{mom}, as it can deal with new sequences incrementally.

In the most extreme case, the experience of a language model is simply the word
sequences that it gets fed and its ``goals'' are to maximize the probability of
predicting a given word based on a sequence of input words. If we were to
compare this to the human experience, it would be the same as having a
disembodied human that is deaf, mute, blind, with symbol sequences like words
being directly encoded into the brain with some form of intrinsic motivation to
produce the symbols that would be more likely to occur in any given instance.

Misinterpreting what \textit{representations}, or rather, what the model states
are capturing in these models leads to confusion about how these models could be
\textit{understanding} symbols, and how this differs from human understanding
\cite{benderClimbingNLUMeaning2020}. Because structural patterns found in
language and word usage reflect how we talk about the world, we tend to ascribe
to these representations the capacity to capture what we could call
\textit{meaning}. This same confusion emerges in the literature in the form of
tasks such as ``semantic similarity'' tasks. The goal of a semantic similarity
task is as follows. Given a set of symbols such as words, measure how similar
their meanings are. In this context meaning, what we mean is a vague human
judgment of similarity, datasets on semantic similarity are built by asking
humans to rate the similarity of words, usually in the context of other words.
The work in \cite{mikolov2013linguistic} for example aims to show how word
vector representations (often called \textit{embeddings}) captured semantic
regularities by using the vector offset method to answer. While this
outperformed the best systems at the time in that particular task, the fact was
that some analogies were encoded in the geometry of the model
\textit{embeddings} --for example king $\rightarrow$ man as queen $\rightarrow$
woman. This means that some patterns of use lead to that particular structure in
those models, particularly when the model was trained with objectives such as
maximizing the likelihood of predictions of missing words in a close task: where
certain words are removed from their context and the model must correctly
predict the missing elements.

Current state-of-the-art work on neural network language models differs only in
a few aspects from the previously cited work such as \cite{bengio2000neural}.
More precisely, current models make use of different inductive biases such as
attention mechanisms, different objective functions, and larger datasets
\cite{vaswani2017attention}. It is understandable that mechanisms that learn to
parametrize the relationships between multiple elements in a sentence, and weigh
them accordingly to make decisions about what words to predict make some of
these models more performant in language modelling. 

Recent work showed that a combination of structural properties and large amounts
of data is enough to produce sequences of words that can be said to be almost
indistinguishable from something produced by a human. Moreover, using human
feedback (as opposed to just word prediction from a given context) is
fundamental to account for our expectations of how language ought to be produced
when interacting with humans \cite{ouyang2022training}. This does not imply
however that any form of human understanding is achieved in the process. As an
analogy, if one learns to bark at a dog to make it behave in a certain way if
the desired behaviour is observed, we can't claim to have mastered \textit{dog
speech}. Similarly, a machine can learn to produce language when engaging with
humans as a means to achieve its goals, but the symbols it uses in the process
are only meaningful in the light of its goals, grounded only in its experience,
not in the same kind of human experience where the original norms are born. As
Wittgenstein famously remarked, ``if a lion could speak, we could not understand
him'' \cite{wittgenstein2010philosophical}.

Computational language models are capable of using multiple objectives/goals
(e.g. human feedback \cite{ouyang2022training}) when learning representations
for linguistic symbols they process. Moreover, such models are shaped not only
by shallow structural patterns in language, but also by other modalities such as
images, or sensors/feedback from robotics tasks \cite{driess2023palm}, makes
these connectionist approaches a good starting point as candidates for
\Gls{mom}. We believe there are two major obstacles to their full potential as
\gls{mom}s coexisting with humans in our linguistic worlds: (1) input symbols
are treated as members of a fixed lexicon; (2) language models are trained to
output and update full probability distributions.

\subsection{The Prediction Frame Problem}
Social media features the use of what is often called ``bad language'': text
that defies our expectations about vocabulary, spelling, and syntax. The
\Gls{nlp} community's response to this phenomenon has largely followed two
paths: normalization and domain adaptation. By adopting a model of
normalization, we declare one version of the language to be the norm, and all
others to be outside that norm; by resorting to domain adaptation we can improve
accuracy on average for tasks that deal with that domain, but it is certain to
leave many forms of language out, because certain niches are not necessarily
coherent domains \cite{eisenstein2013bad}. Both problems expose a fundamental
methodological issue in computational approaches to language, the fact that
models cannot adapt to unseen lexicon. The question becomes: how do we learn to
model or make predictions in domains where a lexicon is not available a priori
(e.g., how do we train classification models on an unknown number of classes)?

Neural language models have found success in many prediction problems from
machine translation to speech recognition, and even in ill-defined and
open-ended tasks like answering questions in conversational settings. Part of
what enables this success is how the problem space is encoded and the learning
procedure that allows for the implicit generalization for unseen data. Neural
language models map discrete representations like word strings into distributed
(vectorial) representations in a geometric space. The output of a language model
is often a discrete probability distribution of words given an input context (a
set or ordered sequences of words). A neural network model can only deal with a
finite lexicon. For example, if a model predicts the occurrence of a symbol in a
sequence, it must do so in light of the probabilities of all other possible
symbols. The architectures of neural language models are tied to how we encode
the lexicon to be used and vocabulary sizes affect these architectures in such a
way that large vocabularies often lead to models that are intractable both to
train and to do inference.

In the past, to make these models feasible one would take the most frequent
\textit{content} words that did not include \textit{function} words such as
\textit{the}, \textit{of}, \textit{a}, \textit{o} and map any other word seen to
a special \textit{UNKNOWN} token. More recent work reduces the vocabulary
through the use of techniques like byte-pair encoding (BPE) \cite{gage1994bpe},
where the most frequent words are encoded as single tokens, while the less
frequent words are composed of multiple (more frequent) tokens, each of them
representing a word part. In other words, the vocabulary of these models is
transformed from words to sub-words symbols based on the frequency of their
usage. The problem is that it still makes it impossible for these models to
learn about unseen words unless they can be encoded as a sequence of other known
subword elements. If the input lexicon changes drastically from the
pre-established lexicon, the only solutions are either to retrain the models or
to make them so large that they encompass a wide variety of possible words or
subwords for data that might appear in the future.

While the average English speaker has a vocabulary in the range of $10,000$ to
$20,000$ words \cite{zechmeister1995growth, brysbaert2016many}, it is not
uncommon that a large language model has vocabularies with $30,000$ to
$100,000+$-symbols \cite{touvron2023llama}, even after reducing the vocabulary
size with the mentioned techniques. That is also considering that most large
language models are trained on uniform languages and don't take into account
code-switching: linguistic behaviour that juxtaposes passages of speech
belonging to two different grammatical systems or sub-systems, within the same
exchange. For example, in Hong Kong, there is code-switching between two
grammatical systems: Cantonese and English; with Cantonese as the dominant
language, contributing bound morphemes, content and function words, and English
being the ``embedded language'' and contributing with lexical, phrases or
compound words \cite{li2000cantonese}.

From an engineering perspective, decomposition in characters or subword tokens
can be a solution to deal with an infinity of concepts and dynamic domains, but
it does not present a solution to the fundamental issue of building \gls{mom}
capable of creating the semantics for whatever symbols they might encounter in
the world. Decomposition works as a frame for tasks that involve human language
but much like the original frame problem, the restrictions on which atoms are
chosen for the decomposition are up to the designer. Fundamentally, it does not
answer the question of how to learn the semantics of signs in unbound domains. 

There is a lot to be said about subword decomposition in the semantics of human
language, especially in languages where morphology plays an important role in
how words are derived and perceived. In humans, understanding the morphological
relationship between words seems to be correlated with reading aptitude even in
different languages \cite{ku2003morphological}. Moreover, when children are
learning how to read, they also start displaying the ability to connect sub-word
patterns to word semantics. This implies that being aware of morphological
features might lead to more efficient models and, perhaps more critically, it
can help with learning models for low-resource languages by generalising
morphological features from high-resource languages
\cite{chaudhary2018adapting}. Nevertheless, the goal when building
\textit{machines of meaning} is not solely to model existing language, but to
create systems capable of acquiring semantics for any system of symbols whether
this system is established or emerging, for example from an artificial language
developed by computational agents \cite{steels1995self}. We are faced with the
following questions:

\begin{itemize}
    \item (1) How to build computational systems to develop representations
    supporting infinitely many concepts from an unspecified lexicon acquired from experience?

    \item (2) How to model language in a context where all the possible outputs
    for a symbol distribution are unspecified and possibly infinite? 
\end{itemize}

Question (1) is not only key for solving the problem of creating artificial
machines of meaning, closer to their human counterparts, but also intrinsically
connected to the problem of continuous learning in \gls{ml}. Large neural
language model parameters are usually trained, and their parameters are frozen
afterwards. The lack of adaptation over time means the input distribution drifts
without the models adapting to these changes and so modelling performance
degrades over time. Another consequence of this lack of adaptation, particularly
prominent in large language modes is that training data contamination leads to
overemphasized capabilities of the models on unseen tasks (zero-shot, or
few-shot learning). Model capabilities are explained by the fact that the tasks
or prompts made to language models have been seen before in training data and
not by a model capacity to generalize to unseen data \cite{LiTask2023}.

Question (2) is an intrinsic extension of (1), but also a form of a
\textit{frame problem}: how do we build machines that make predictions about
categories without considering all the categories that can be conceived, or even all other categories the model uses to make predictions?

There are two ways in which fixed lexicons restrict neural network architectures
(preventing us from learning incremental models for possibly infinite
vocabularies). Firstly, \textit{representing} discrete entities, and converting
them into distributed representations requires the network inputs to be fixed in
accordance to the cardinality of an existing vocabulary. Secondly, in order to
model language, most implementations are built to output predictions for words
or symbols based on a sequence of previously observed words. These predictions
have the form of probability distributions. This means the output dimensions in
a network are determined by the number of words in a pre-established vocabulary,
and by extension, the number of parameters in the said networks is tied to
vocabulary size.

We believe that a solution to create true \Gls{mom} lies primarily in overcoming
these two restrictions. The restriction on the number of symbols that can be
processed by language models (i.e. the conversion of discrete symbols to
distributed representations) could be tackled with compressive encoding using
for instance dimensionality reduction techniques like random projections
\cite{kanerva2009hyper,achlioptas2003database,nunes2018neural}. The main
challenge of adopting new forms of symbol encoding lies in the fact that
virtually all empirically effective \gls{llm} architectures have been designed
with the traditional \textit{1-of-V} encoding where each symbol is mapped to a
respective parameter vector. Changing the encoding would mean that we would need
to find the architectures that would be effective for different encoding scheme.

We consider the second issue a form of frame problem because much like classic
AI, a related issue question arises of how to compute the consequences of an
action without the computation having to range over the action's non-effects.
The solution to the computational aspect of the frame problem adopted in most
symbolic AI programs is some variant of what McDermott calls the “sleeping dog”
strategy \cite{hanks1987nonmonotonic}. The idea here is that not every part of
the data structure representing an ongoing situation needs to be examined when
it is updated to reflect a change in the world. Rather, those parts that
represent facets of the world that have changed are modified, and the rest is
simply left as it is (following the dictum “let sleeping dogs lie”). 

The solution to the second issue implies we would have to do away with
approaches like \Gls{mle}, usually employed with probabilistic neural language
models \cite{bengio2000neural}. This remains computationally prohibitive, and,
despite the advances in hardware making it possible for large language models to
be trained (e.g. \cite{touvron2023llama}), it has become clear that large neural network models have become unsustainable both from a scalability perspective and an environmental one.

The solution to this issue might be found in approaches like likelihood-free
estimation where the focus is in the solution of problems where obtaining the
gradients for the log-likelihood of categories is intractable (such as the
contrastive divergence method proposed in \cite{hinton2002contrastive}). For
example, we can look at neural language models as instances of non-normalized
statistical models, but despite the research into techniques designed to train
such models like \textit{Score Matching} \cite{hyvarinen2005estimation}, or
\textit{Noise Contrastive Estimation (NCE)}\cite{gutmann2010noise}, these approaches have remained largely under-explored in current implementations of \Gls{llms}.

In summary, the combination of likelihood-free methods and compression encoding
schemes like sparse random representations
\cite{kanerva2000random,nunes2018neural} can potentially enable the creation of
models that learn incrementally, and reuse parameters without catastrophic
forgetting. These could, in principle, become the foundations of what we can
consider to be \Gls{mom}, but the challenges of inference posed by this new way
of looking at sequence modelling demand further research to make them a
competitive alternative to today's implementations of \gls{llm}s. 

%TODO sketch a solution

\section{Conclusion}

In any \textit{machine of meaning}, semantics should be established based on the
machine's goals and motivations. If we wish to design such machines closer to
our idealized (but often ill-defined) human capabilities, they must also not
fall prey to the modern \textit{frame problem}: they must be capable of
processing and incorporating any given symbols as they appear in the context of
their use. Ignoring this problem leads to both computational tractability
problems and methodological limitations. It has become clear that these issues
lead to computational models that are unsustainable, both from a scalability
perspective (with issues of modern neural network architectures being a common
subject of research \cite{Feng2024}), and now more than ever, from an
environmental perspective, with energy usage far outpacing hardware efficiency
gains (e.g. for large scale neural network models \cite{Strubell2019,
benderDangersStochasticParrots2021, BolonCanedo2024}). 

Current approaches to solving language tasks are limited to the kinds of
inductive bias we can incorporate into models; models are limited to
applications to well-defined static environments where we trade off massive
computational (and energetic) costs for the capacity to create systems that
appear, eloquent syntax-following language producers, while still being limited
to the data on which they were originally trained; incremental online learning
is not possible since what symbols the current models are capable of processing
are tied to their respective architectures. 

We clarified both the scientific challenges and the philosophical problems
surrounding the notions of \textit{meaning} and \textit{grounding}, and how
these connect to linguistic agent motivations and language use. Moreover, we
highlighted how ill-equipped the current computational approaches such as
\Gls{llms} are to deal with human language. Such approaches are not only lacking
proper grounding mechanisms for language semantics, but also limited by their
architectures in such a way that is not possible to adapt to an open and dynamic
linguistic domain. Our analysis provides starting point for new discussions,
avenues of research, and solutions to the highlighted problems to create and
understand all forms of \Gls{mom} beyond anthropocentric language use.    

To create \textit{machines of meaning}, we must be able to deal with problems
with no a priori boundary to what information is relevant to a given process. In
the classic symbolic approaches to AI, this was called the frame problem, in the
current iteration we call this the prediction frame problem. It is clear that
incremental learning approaches are key for adapting to the open nature that the
language domain display. Meaning, is not just born out of language use, but from
the connection between language use, a non-enumerable number of possible
contexts for its use, and the ever-evolving intrinsic goals of each linguistic
agent (human or otherwise) that motivates its use.

\bibliography{main.bib}

\end{document}